\documentclass[conference]{IEEEtran}
\IEEEoverridecommandlockouts                                     
\usepackage{cite}
\usepackage{amsmath,amssymb,amsfonts}
\usepackage{algorithmic}
\usepackage{graphicx}
\usepackage{textcomp}
\usepackage{booktabs} 
\usepackage{subcaption}
\usepackage{diagbox}
\usepackage{url} 
\usepackage{hyperref}
\hypersetup{colorlinks,allcolors=blue}
\usepackage{fancyhdr}
\usepackage{xcolor}
\usepackage{adjustbox} 
\usepackage{graphicx}
\usepackage[inline]{enumitem}
\usepackage{multirow}
\def\BibTeX{{\rm B\kern-.05em{\sc i\kern-.025em b}\kern-.08em
    T\kern-.1667em\lower.7ex\hbox{E}\kern-.125emX}}
\makeatletter
\newcommand{\linebreakand}{%
  \end{@IEEEauthorhalign}
  \hfill\mbox{}\par
  \mbox{}\hfill\begin{@IEEEauthorhalign}
}
\makeatother

\begin{document}

\title{Human Stone Toolmaking Action Grammar (HSTAG): A Challenging Benchmark for Fine-grained Motor Behavior Recognition
}
\author{Cheng Liu$^{1,*}$, 
Xuyang Yan$^{2}$,
Zekun Zhang$^{3}$,
Cheng Ding$^{4}$,
Tianhao Zhao$^{4}$,
Shaya Jannati$^{1}$,\\
Cynthia Martinez$^{1}$,
and Dietrich Stout$^{1}$ 
\thanks{$^{1}$Department of Anthropology, Emory University, Atlanta, Georgia, USA.
Emails: {\tt cheng.liu@emory.edu; shaya.jannati@emory.edu; cynthia.martinez2@emory.edu; dwstout@emory.edu}}
\thanks{$^{2}$Department of Electrical and Computer Engineering, North Carolina A\&T State University, Greensboro, North Carolina, USA. 
Email: 
{\tt xyan@aggies.ncat.edu}}%
\thanks{$^{3}$ Department of Computer Science, Stony Brook University, Stony Brook, New York, USA.
Email: {\tt zekzhang@cs.stonybrook.edu}}%
\thanks{$^{4}$Department of Economics, Emory University, Atlanta, Georgia, USA. 
Emails: {\tt cheng.ding.emoryecon@gmail.com; tianhao.zhao@emory.edu}}%
\thanks{$^{*}$Corresponding Author}%
}

\maketitle

\begin{abstract}
Action recognition has witnessed the development of a growing number of novel algorithms and datasets in the past decade. However, the majority of public benchmarks were constructed around activities of daily living and annotated at a rather coarse-grained level, which lacks diversity in domain-specific datasets, especially for rarely seen domains. In this paper, we introduced Human Stone Toolmaking Action Grammar (HSTAG), a meticulously annotated video dataset showcasing previously undocumented stone toolmaking behaviors, which can be used for investigating the applications of advanced artificial intelligence techniques in understanding a rapid succession of complex interactions between
two hand-held objects. HSTAG consists of 18,739 video clips that record 4.5 hours of experts' activities in stone toolmaking. Its unique features include (i) brief action durations and frequent transitions, mirroring the rapid changes inherent in many motor behaviors; (ii) multiple angles of view and switches among multiple tools, increasing intra-class variability; (iii) unbalanced class distributions and high similarity among different action sequences, adding difficulty in capturing distinct patterns for each action. Several mainstream action recognition models are used to conduct experimental analysis, which showcases the challenges and uniqueness of HSTAG \href{https://nyu.databrary.org/volume/1697}{\nolinkurl{https://nyu.databrary.org/volume/1697}}.
\end{abstract}


\section{Introduction}
\label{sec:intro}

Serial ordering of human behaviors in stone toolmaking activities has been a challenging scientific problem in understanding a rapid succession of complex interactions between two hand-held objects \cite{stoutStoneToolmakingEvolution2011}. It aims to capture the distinct characteristics of those sophisticated hand actions from the visual representations with a well-defined linguistic syntax or action grammar, which refers to the rules and structures governing the sequence and composition of actions, which is crucial for uncovering the evolutionary history of this ancient human technology. This scientific problem can be formulated as an action classification task and appears as a practical example of an ``AI for science" problem. The investigation of this problem helps to uncover the story behind the evolutionary history of the most widespread human technological activity, which brings a lot of significance to social science research. Yet, limited studies and efforts have been devoted to studying the potential of the advanced artificial intelligence (AI)-based approaches \cite{zhang2024luwa}. 

Beyond the impact in social science, several unique challenges of this visual-based action understanding problem are posed to the computer vision (CV) community versus other action classification problems. Firstly, the fast transition among high-frequency actions and high frame-level similarity among different action sequences increase the difficulty in discovering key feature representations for each action class. Secondly, hand actions are usually captured as small regions within each frame, and the noises caused by the variety \& complexity of background regions will confuse the learning procedure, which can degrade the accuracy of modern CV approaches. Thirdly, the sample distribution of each action class is highly unbalanced and thus increases the chance of obtaining a poor feature representation for the underrepresented action classes.

The recent advancements in CV approaches have been extensively investigated for human action recognition tasks, and substantial progress has been made in \cite{xia2020, wangSurveyDeepLearningbased2023, liuSurveyVideoMoment2023, bhoiSpatiotemporalActionRecognition2019}. More than 30 video-centered action recognition datasets \cite{soomroUCF101Dataset1012012, kayKineticsHumanAction2017, kuehne2011, heilbron2015, andriluka2014, sigurdsson2016, idrees2017} were introduced as the benchmarks for the validation and evaluation of novel effective CV techniques. Despite the presence of numerous action recognition datasets, the lack of diversity in domain-specific datasets limits the investigation of the state-of-the-art (SOTA) advanced CV approaches in benefiting rarely seen domains especially for stone toolmaking. Addressing this gap requires the inclusion and establishment of high-quality domain-specific datasets. However, the creation and curation of domain-specific datasets not only requires massive expert knowledge but also high labor costs for data annotations, which prevents the collection of a well-represented dataset that can reflect the variability of this domain.  

In light of the challenges and opportunities described above, we introduced an open-source video dataset (HSTAG) with fine-grained annotation information for stone toolmaking activity understanding. It consists of $18,739$ video clips that record $4.5$ hours of experts' activities in stone toolmaking. Several major characteristics of HSTAG include: (i). \textbf{Multiple angles of view.} Each action class consists of video clips recorded from either the top view or front view, and it increases the inner-class variability on the patterns; (ii). \textbf{Switch among multiple tools.} Different types of tools, such as stone and antler, are used for the same action, which can decrease the within-class similarity; (iii). \textbf{Short action sequence and high-frequency action transitions.} Some actions last for short time periods, and the transition between different actions is fast, which makes it hard to capture the primary characteristics. The introduced HSTAG dataset aims to benefit CV-related research from two aspects: firstly, enrich the domain-specific benchmark for both the performance evaluation of SOTA CV techniques; secondly, drive the development of novel CV approaches on classifying temporal action sequence with unbalanced class distributions.


Overall, the main contributions of our work are twofold: 1) We introduce the first open-source video dataset (\textbf{HSTAG}) with fine-grained annotations for action recognition tasks, and it includes $11,778$ videos that display a novel set of behaviors from the rare domain, namely stone toolmaking. HSTAG distinguishes itself from other popular action recognition datasets with its extremely short action duration and frequent action transitions, making itself a challenging benchmark for new action recognition algorithms development and testing. 2) We evaluated the generalization capability of several mainstream video classification algorithms (VideoMAEv2 \cite{wang2023videomaev2}, TimeSformer \cite{bertasius2021space}, and ResNet \cite{he2016deep}) on a rarely seen domain (stone toolmaking) and compared a series of key metrics of their performances on other benchmarks. Experimental analysis revealed further challenges, including highly imbalanced action classes and nuanced similarities between certain actions.
\section{Related work}
\label{sec:related-works}


Recent technological advancements, particularly in deep learning, have significantly influenced the development of more sophisticated models for video analysis. The introduction of transformer-based models, such as the TimeSformer by Bertasius et al. \cite{bertasius2021space}, has provided new avenues for capturing spatial-temporal relationships in videos. The Video Swin Transformer by Liu et al. \cite{liu2022video} further extends this innovation, showing notable improvements in action recognition. Additionally, the Co3D model by Reizenstein et al. \cite{reizenstein2021common} applies 3D reconstruction techniques to common objects, enhancing the understanding of complex scenes and activities in videos.

In addition to these technological advancements, there has been an increasing focus on constructing benchmark datasets for action recognition. According to a non-exhaustive survey from the crowd-sourcing platform Papers with Code\footnote{\url{https://paperswithcode.com/datasets?task=action-recognition\&mod=videos\&page=1}}, computer scientists have constructed at least 30 video-centered datasets for the task of action recognition, ranging from large-scale comprehensive ones that contain more than 100 hours of videos \cite{soomroUCF101Dataset1012012, kayKineticsHumanAction2017, kuehne2011, heilbron2015, andriluka2014, sigurdsson2016, idrees2017} to more specialized activities such as cooking \cite{damen2022, kuehne2014}, gymnastics \cite{shao2020}, etc. Nonetheless, many of these existing databases are constructed in a rather coarse-grained manner, meaning the actions composed of multiple sub-actions are not fully parsed to their minimally identifiable units.

Meanwhile, the past few years have also witnessed the growing interests in segmenting specific and intricate human activities. EPIC-Kitchens, introduced by Damen et al. \cite{damen2018scaling}, provides detailed annotations for cooking activities, emphasizing the need for granular action recognition. Further contributing to this trend, the SOMETHING-SOMETHING dataset by Goyal et al. \cite{goyal2017something}, offers a diverse collection of human-object interactions, facilitating the study of fine-grained actions. These detailed datasets are essential for understanding complex tasks like stone toolmaking, where actions are highly specialized and nuanced, underscoring the importance of specificity and detail in dataset construction.

Most recently, a microscopic image dataset with different magnifications and sensing modalities is proposed to facilitate lithic use-wear analysis for reconstructing the function of ancient stone tools \cite{zhang2024luwa}. The authors collaborated with archaeologists to obtain high-quality annotation information on the collected microscopic images using expert knowledge. As the first open-source dataset that focuses on classifying ancient tools, it complements existing image classification benchmarks on common objects. Beyond the object classification tasks, we introduced the HSTAG video dataset to investigate the generalizability of existing advanced AI techniques in understanding human actions in ancient toolmaking.

\section{Human Stone Toolmaking Action Grammar dataset}

Here, we present the details of the Human Stone Toolmaking Action Grammar (\textbf{HSTAG}) dataset.
\begin{table}[t]\centering
\setlength{\tabcolsep}{3pt}
    \begin{tabular}{@{}lcccc@{}}\toprule
    \multirow{2}{*}{Dataset} & Duration & Actions & \# of & Avg. clip \\
                             & (hr) & classes & clips & time (s) \\ \midrule
    Kinetics 400~\cite{kayKineticsHumanAction2017} & $-$ & 400 & $-$  & $-$ \\
    Kinetics 600~\cite{Carreira2018ASN}   & $-$ & 600 & $-$  & $-$\\
    \begin{tabular}{@{}l@{}}Something-\\Something-V2~\cite{goyal2017something}\end{tabular} & $-$ & 174 & 221K & $-$ \\
    UCF-101~\cite{soomroUCF101Dataset1012012}& 26.7& 101 & 13K  & 7.21\\
    UCF-Sports~\cite{Soomro2014,4587727}     & 0.267& 10 & 150  & 6.39\\
    COIN~\cite{COINDATA}                     & 476  & 180& 46K  & 14.91 \\
    HACS~\cite{zhao2019hacs}                 & $-$ & $-$ & 1.6M & 2\\
    FineGym~\cite{shao2020}                  & 708 & 530 & 32K  & 2 \\\midrule
    \textbf{HSTAG} (Ours)                    & 4.5& 7   & 18.7K & 0.91 \\ \bottomrule
    \end{tabular}
\caption{Comparison of dataset statistics. K and M represent $10^3$ and $10^6$ respectively.}
\label{tab-cpchar}
\end{table}

\subsection{Data collection}

The videos of human stone toolmaking used in this project were generated from multiple existing experimental archaeology projects involving stone toolmaking experts, which were reviewed and approved by the Institutional Review Board of Emory University and the Research Ethics Service at the University College London.
All research participants demonstrating stone toolmaking behaviors in our videos are experts who have signed the consent form and are aware of the risk of losing anonymity because their images and/or voices in the recordings may be seen or heard and thereby identified by people who are familiar with them.

\begin{figure}[t]\centering
    \includegraphics[scale=0.5]{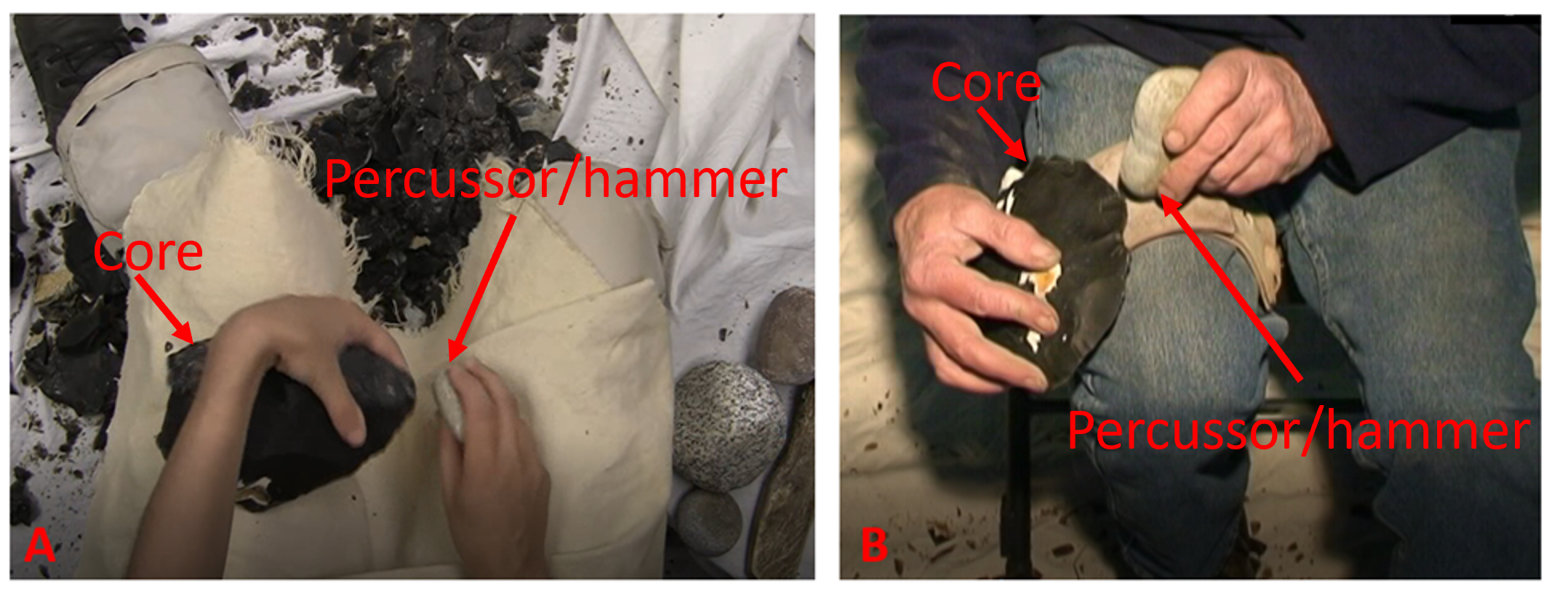}
    \caption{Stone toolmaking shot from the top/ego view (A) and the front view (B).} \label{fig:views}
\end{figure}

We recorded these videos in two different views, including the top/ego view captured by head-mounted cameras (Figure \ref{fig:views}A) and the front view shot through regular video cameras set on tripods and smartphones (Figure \ref{fig:views}B).
Within the front-view videos, a sub-collection focuses solely on the hand movements and the lower body of the research participants, while in other videos, the whole body of research participants, including faces, is displayed.
In total, we recorded 4.5 hours of expert videos and finished the video annotation for all expert videos. We have excluded the last label in each video because it lacks an ending timestamp. On average, a single action in the HSTAG dataset lasts for approximately 0.91 seconds. 

\subsection{Data annotation} 

\begin{figure}[t]\centering
\includegraphics[width=\linewidth]{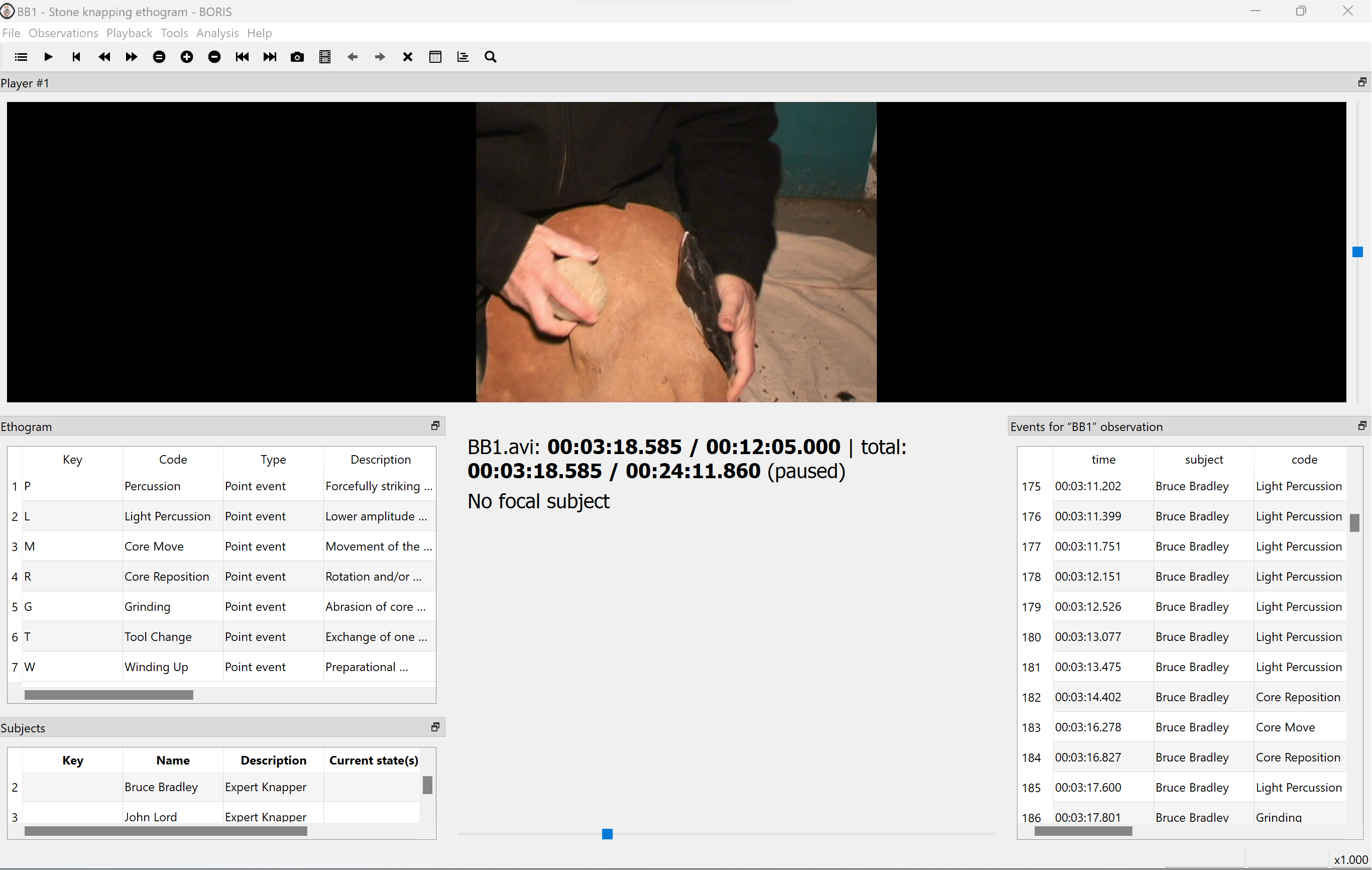}
\caption{The user interface of the BORIS annotation tool.}
\label{fig:BORIS}
\end{figure}

We use BORIS \cite{friard2016}, an open-access software developed for action labeling in videos and audio, as the main platform for data annotation (Figure \ref{fig:BORIS}) .
More specifically, we adopt an ethogram that is known as action grammar and originally developed by Stout et al. \cite{stoutMeasurementEvolutionNeural2021} for the analysis of stone knapping behavior with some minor modifications.
Stone knapping is generally composed of a limited set of actions that are comparable to lexicons in our natural languages, and the order of actions within a certain mode of stone tool technology is somewhat flexible yet still constrained.
To this extent, knapping and other actions can, in certain instances, be likened to syntax.
Action grammars were designed to quantify the complexity of some of the earliest stone tool technologies in human (pre)history (e.g., Oldowan and Acheulean technologies) and detect the possible co-evolutionary relationship between tool making and language evolution, which both are characterized by goal-directed behavior with a certain level of hierarchical structure.

\begin{figure*}[t]\centering
\includegraphics[width=\linewidth]{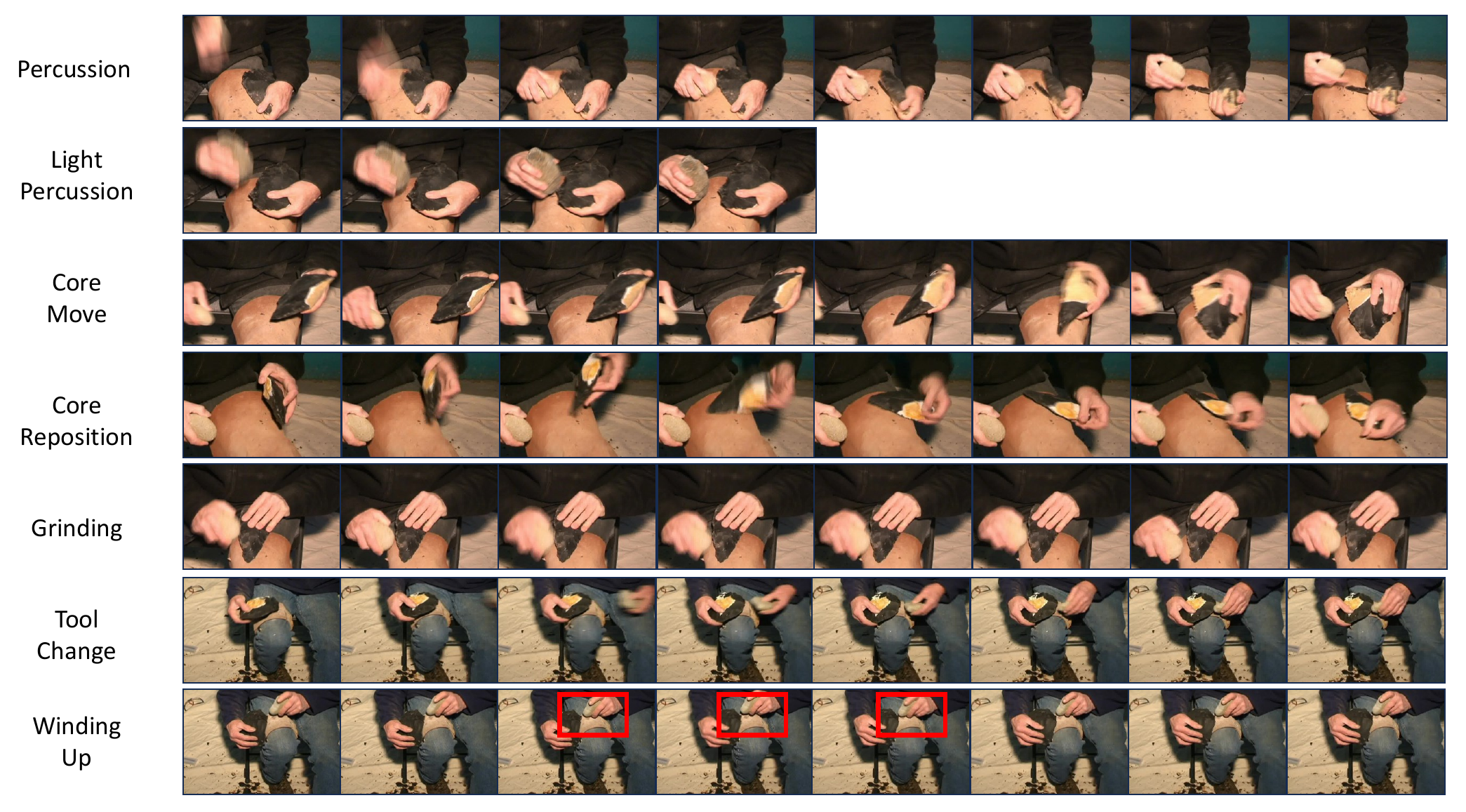}
\caption{Samples for each of the 7 actions in HSTAG. We use red rectangles to highlight the participant's slight hand movement.}
\label{fig-action-illustration}
\end{figure*}

Our modified ethogram \cite{Liu2024} includes the following seven distinct actions: 
\begin{itemize}
    \item \textbf{Percussion}: Forcefully striking core with percussor (hammerstone or antler billet) in such a way as to potentially remove a flake.
    \item \textbf{Light Percussion}: Lower amplitude tangential strike to the tool edge of the kind often employed for platform preparation.
    \item \textbf{Core Move}: Movement of the core without a change in grip, which often occurs during the core inspection.
    \item \textbf{Core Reposition}: Rotation and/or reorientation of the core involving the repositioning of the hand, which is often associated with the transition to a new percussion target.
    \item \textbf{Grinding}: Abrasion of core edges using a hammerstone. The abrasion movement should come from at least two different directions.
    \item \textbf{Tool Change}: Exchange of one percussor for another.
    \item \textbf{Winding Up}: Preparational percussor movements towards the core that do not lead to the detachment of flakes, which can either be in direct contact with cores or not.
\end{itemize}

Figure \ref{fig-action-illustration} displays the sequence of actions for all seven action classes above. From Figure \ref{fig-action-illustration}, several important observations can be obtained as follows: (i). different action classes share the similar sequence of actions; (ii). the presence of background noises decreases the dissimilarity between action classes. These characteristics significantly increase the difficulty in learning the distinct representation for each action class, which partially constitutes the challenging aspects of HSTAG.

Two trained annotators participate in the annotation procedure.
With these two annotators, a domain-specific expert coordinates weekly-based cross-validation of labeled video segments to increase the inter-observer reliability. 
During the cross-validation procedure, we align the labeling results until they reach a complete agreement in terms of the total label number and the content of each individual label.
We also conducted further calibration to make sure that the starting point time difference between the two sets of annotations is within one second, which is the most time-consuming step. We make both annotators' results public.

All output files are in the form of Excel sheets, which is one of the default formats of the BORIS software. Each Excel sheet consists of two parts. The first part (Row1 - Row15) contains the meta-information regarding the video coded, including the file name, path, labeling time, etc. The second part (Row16 to the last row) represents the coded action sequences with 7 variables that are particularly relevant here. The first variable is time. All these actions are stated event behavior with a starting time and ending time. However, we treated them as point event behavior with only one-time stamps for labeling simplicity.
This is because actions like light percussion feature relatively high velocity, and sometimes the software can not effectively detect the trivial time difference between the starting point and ending point even when the video is played at $\times$0.3 speed.
Therefore we only created the time stamp for the starting point of an action, and its ending point is treated as the starting point of the subsequent action. 

The total length is the length of the video in seconds. The subject contains the information of the research participants, among which experts' real names are provided. Behavior is our classification of the behavior, which must be one of the seven actions defined above. The behavior category has two options: DH (Dominant hand) or SH (Supporting hand), which are used to characterize the main carrier of the given action. Only a small portion of our participants are left-handed, and the handedness information is available in the main data sheet. Modifier 1 represents two groups of modifiers that are respectively associated with the action of ``Tool change'' and ``Winding up''. The modifiers for ``Tool change'' include HH (Hard Hammer, usually round stones) and SH (Soft Hammer, usually antler), indicating the physical property of the new percussor picked by the research participant. Since these two types of percussion have distinct morphology, we thought it might be crucial for machine learning algorithms to learn this piece of information. The modifiers for ``Winding up'' include contact and no contact, indicating if the percussor physically touched the core or not. The last variable is Status, and as mentioned above all actions are treated as point event action. 

\subsection{Dataset statistics}

To accurately capture the characteristics of each action, we clipped the collected videos into smaller videos with reference to the timestamp of each action label. As a result, a total number of $18,739$ clipped videos are obtained, and each video describes a particular point-event action. We conducted the shuffling splits to generate the training, and testing sets. With a split ratio of ($75:25$), the training set consists of $14,061$ clipped videos, and the testing set has $4,678$ clipped videos. For each behavior class, we performed uniform sampling in proportion to the size of each class. The class distributions of the obtained training, and testing subsets are presented in Table \ref{table-dataset-stats}. Table \ref{table-dataset-stats} indicates the imbalanced class distributions among all seven action classes, which contributes to the third challenge of the HSTAG dataset. 

\begin{table}[t]
\setlength{\tabcolsep}{3pt}
\begin{tabular}{lrrrrccc}\toprule
\multirow{2}{*}{Action} & \multirow{2}{*}{Percentage} & \multicolumn{3}{c}{\# of instances} & \multicolumn{3}{c}{Duration (s)}\\\cmidrule(lr){3-5}\cmidrule(lr){6-8}
                && Total & Train & Test & Avg. & Mdn. & Std. \\\midrule
Percussion      &11.37\%& 2,130  &1,585& 545& 0.74 & 0.65 & 0.40 \\
Light Percussion&58.24\%& 10,913 &8,215 &2,698& 0.36 & 0.30 & 0.25 \\
Core Move       &12.11\%& 2,270  &1,713 & 557 & 1.28 & 1.00 & 1.24 \\ 
Core Reposition & 9.61\%& 1,800  &1,319 & 481 & 3.22 & 2.16 & 3.67 \\
Grinding        & 0.99\%& 187    & 137  &  50 & 2.71 & 2.16 & 2.05 \\
Tool Change     & 1.96\%& 368    & 284  &  84 & 1.34 & 0.82 & 3.27 \\
Winding Up      & 5.72\%& 1,071  & 808  & 263 & 0.80 & 0.50 & 0.88 \\\midrule
Total           &  100\%&18,739  &14,061&4,678&-&-&-\\\bottomrule
\end{tabular}
\caption{Statistics of the HSTAG dataset. Avg., Mdn, and Std. stand for average, median, and standard deviation, respectively. There is significant imbalance among different action classes. The dataset is split into training/testing sets using the ratio $75:25$. (Note: - refers to not applicable)} \label{table-dataset-stats}
\end{table}


\section{Experiments analysis}
In this section, we conducted the experimental analysis and comparison studies using several SOTA video classification models. Details of the experimental results, as well as several important observations, are discussed here.

\subsection{Benchmark models}
For analysis and comparison studies on the collected video dataset, we test three state-of-the-art (SOTA) video classification models: VideoMAEv2 \cite{wang2023videomaev2}, Time-Space Transformer (TimeSformer) \cite{bertasius2021space}, and ResNet \cite{he2016deep} with Gated Recurrent Unit (GRU) \cite{cho2014rnn} (ResNet+GRU). VideoMAEv2 has shown outstanding performance in learning the high-level representation of redundant video contents, especially in small datasets. TimeSformer model inherits the backbone from the standard Transformer and enables the spatiotemporal feature learning capability, which can effectively improve the learning performance in action recognition. ResNet backbone extracts rich features from each video frame, and GRU module models the temporal correlation among those frames.

\subsection{Experimental setup}
For VideoMAEv2, we use the ViT-small variant, which is distilled from the ViT-giant VideoMAEv2 variant. The ViT-giant variant is pre-trained on Kinetics-710~\cite{Carreira2018ASN} dataset. For the TimeSformer model, we used the model weights pre-trained on the Kinetics-600~\cite{Carreira2018ASN} dataset. For the ResNet+GRU model, we use the ResNet-18 variant pre-trained on ImageNet-1K~\cite{DenDon09Imagenet}. We extract image-level features from each input frame and feed the sequence of 512-dimensional feature vectors to a GRU layer, followed by a fully-connected layer for classification. For VideoMAEv2 and TimeSformer, we replace the last classification layer in the pre-trained model to accommodate the number of classes for HSTAG, and the parameters are randomly initialized before training. For ResNet+GRU, both the GRU and classification layer are randomly initialized. The number of frames in the input video and the spatial resolution of the input frames of each model are shown in Table \ref{table-model-stats}. The models are of different scales, and the number of parameters and giga floating-point operations (GFLOPs) of them are also shown in Table \ref{table-model-stats}.

\begin{table}[t]\centering
\setlength{\tabcolsep}{3pt}
\begin{tabular}{ccccc}\toprule

\multirow{2}{*}{Model} & \multicolumn{2}{c}{Input} & \multirow{2}{*}{Parameters} & \multirow{2}{*}{GFLOPs} \\\cmidrule(lr){2-3}
         & Frames & Resolution && \\ \midrule
VideoMAEv2~\cite{wang2023videomaev2}    &16&$224\times 224$ & 21.9M & 68.5 \\
TimeSformer~\cite{bertasius2021space}   & 8&$224\times 224$ &121.3M & 380.0 \\
ResNet+GRU~\cite{he2016deep,cho2014rnn} & 8&$112\times 112$ & 12.8M & 7.8\\\bottomrule
\end{tabular}
\caption{Temporal and spatial resolution of the input frames, number of parameters, and giga floating-point operations (GFLOPs) of the models. M represents $10^6$.} \label{table-model-stats}
\end{table}

The input video frames are first resampled temporally using the nearest neighbor, according to the number of frames each of the models takes. Then, each frame is resized to the desired spatial resolution and normalized according to the corresponding mean and variance of the pixel values of the pre-training datasets. We apply random cropping as data augmentation. We use cross-entropy loss and Adam~\cite{Kingma2014AdamAM} optimizer to train the models. All the models are trained using the same schedule, which has 20,000 iterations with batch size 8 and learning rate $8\times 10^{-5}$. In the first 5,000 iterations, the pre-trained parts of each of the models are fixed for the models to warm up. The whole network is trained for the remaining iterations. The learning rate is decayed by a factor of 0.1 at iteration 16,666.

\subsection{Evaluation metrics}
We report the test set performance of all three models in terms of classification precision and $F_1$ score \cite{grandini2020metrics}. For each action class, we use the one-\textit{vs.}-rest rule to calculate binary precision and recall score. Then, we used the macro-average to obtain the overall precision and $F_1$ score across all action classes to account for the presence of imbalanced classes. In addition, we measure the multi-class classification accuracy. We do not measure the per-class accuracy since it is not very meaningful in an imbalanced dataset. Among two sets of annotations, we select the annotations from the second annotator as the ground truth.

\subsection{Result discussions}

\begin{table}[t]\centering
\setlength{\tabcolsep}{2.3pt}
\begin{tabular}{clrcccc}\toprule
\multirow{2}{*}{Model} & \multirow{2}{*}{Action} & \multirow{2}{*}{Percentage} & \multicolumn{2}{c}{Class-specific} & \multicolumn{2}{c}{Overall}\\ \cmidrule(lr){4-5} \cmidrule(lr){6-7}
 &&& Pr.$\uparrow$ & Re.$\uparrow$ & Acc.$\uparrow$ & $F_1\uparrow$ \\\midrule
\multirow{8}{*}{\begin{tabular}{c}VideoMAEv2\\\cite{wang2023videomaev2}\end{tabular}} %
 & Percussion       & 15.73\% & 0.808 & 0.855 & - & -\\
 & Light Percussion & 51.52\% & 0.923 & 0.956 & - & -\\
 & Core Move        & 14.72\% & 0.803 & 0.711 & - & -\\
 & Core Reposition  & 11.83\% & 0.796 & 0.830 & - & -\\
 & Grinding         &  0.87\% & 0.829 & 0.580 & - & -\\
 & Tool Change      &  1.30\% & 0.692 & 0.429 & - & -\\
 & Winding Up       &  4.04\% & 0.846 & 0.730 & - & -\\\cmidrule(lr){2-7}
 & Overall          &   100\% & - & - & 0.876 & 0.762  \\\midrule
\multirow{8}{*}{\begin{tabular}{c}TimeSformer\\\cite{bertasius2021space}\end{tabular}} %
 & Percussion       & 15.73\% & 0.776 & 0.807 & -  & -\\
 & Light Percussion & 51.52\% & 0.896 & 0.954 & -  & -\\
 & Core Move        & 14.72\% & 0.779 & 0.607 & -  & -\\
 & Core Reposition  & 11.83\% & 0.722 & 0.794 & -  & -\\
 & Grinding         &  0.87\% & 0.750 & 0.360 & -  & -\\
 & Tool Change      &  1.30\% & 0.642 & 0.405 & -  & -\\
 & Winding Up       &  4.04\% & 0.843 & 0.635 & -  & -\\\cmidrule(lr){2-7}
 & Overall          &   100\% & - & - & 0.845 & 0.695  \\\midrule
\multirow{8}{*}{\begin{tabular}{c}ResNet+GRU\\\cite{he2016deep,cho2014rnn}\end{tabular}} %
 & Percussion       & 15.73\% & 0.793 & 0.761 & -  & -\\
 & Light Percussion & 51.52\% & 0.878 & 0.963 & -  & -\\
 & Core Move        & 14.72\% & 0.749 & 0.648 & -  & - \\
 & Core Reposition  & 11.83\% & 0.768 & 0.688 & -  & -\\
 & Grinding         &  0.87\% & 0.643 & 0.180 & -  & -\\
 & Tool Change      &  1.30\% & 0.571 & 0.286 & -  & -\\
 & Winding Up       &  4.04\% & 0.785 & 0.681 & -  & -\\\cmidrule(lr){2-7}
 & Overall          &   100\% & - & - & 0.837 & 0.644  \\\bottomrule
\end{tabular}
\caption{Model performance in terms of class-specific precision (Pr.), recall (Re.), overall accuracy (Acc.), and macro-averaged F1-score ($F_1$). The class-specific accuracy is not meaningful on an imbalanced dataset thus not shown. The models are trained on HSTAG  and the percentage column shows the percentage of the instances of each action in the test split. (Note: - refers to not applicable)}\label{table-res}
\end{table}

\begin{figure*}[t]
\begin{subfigure}{0.33\linewidth}
\includegraphics[width=\linewidth]{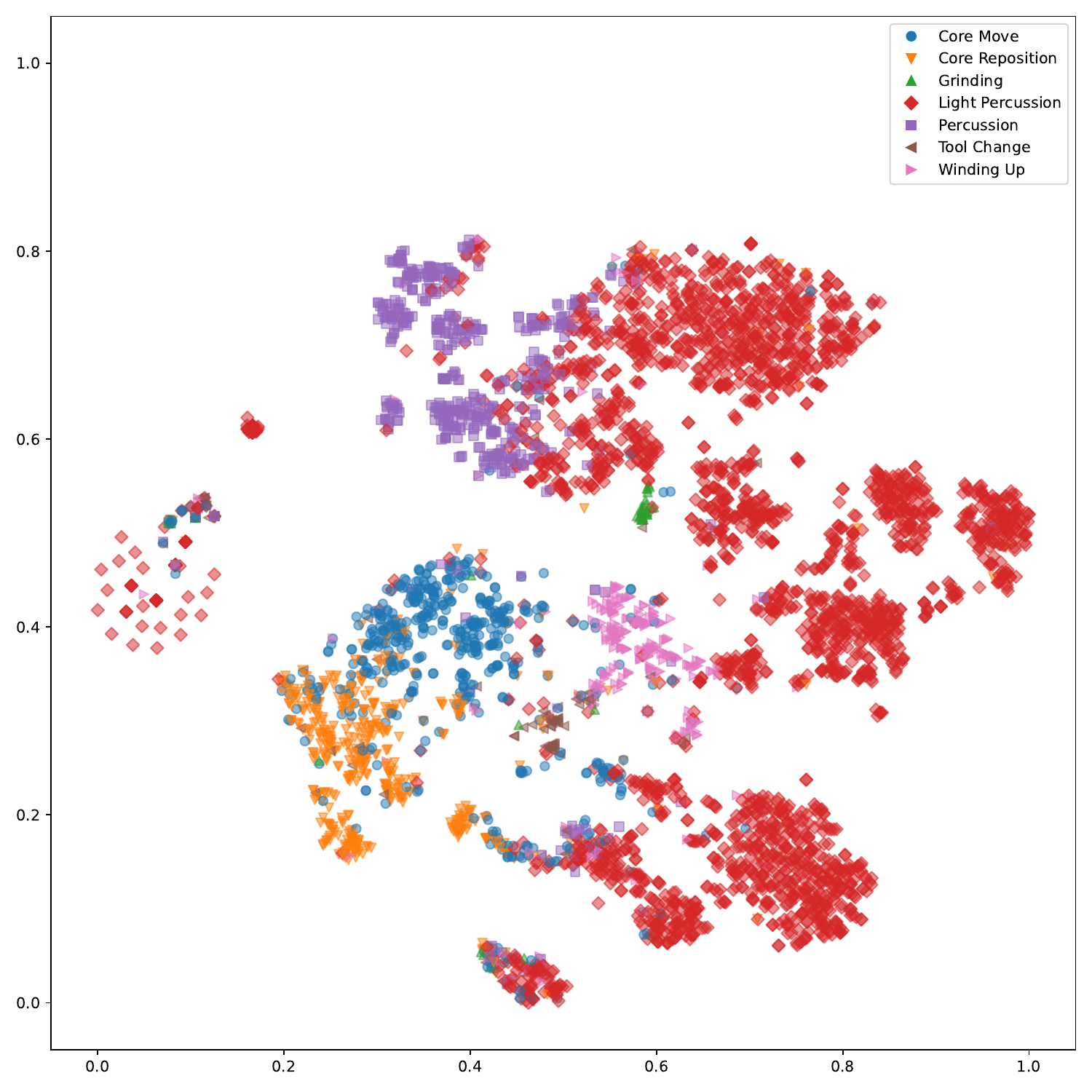}
\caption{VideoMAEv2 model.}\label{tab:visual-videomae}
\end{subfigure}
\begin{subfigure}{0.33\linewidth}
\includegraphics[width=\linewidth]{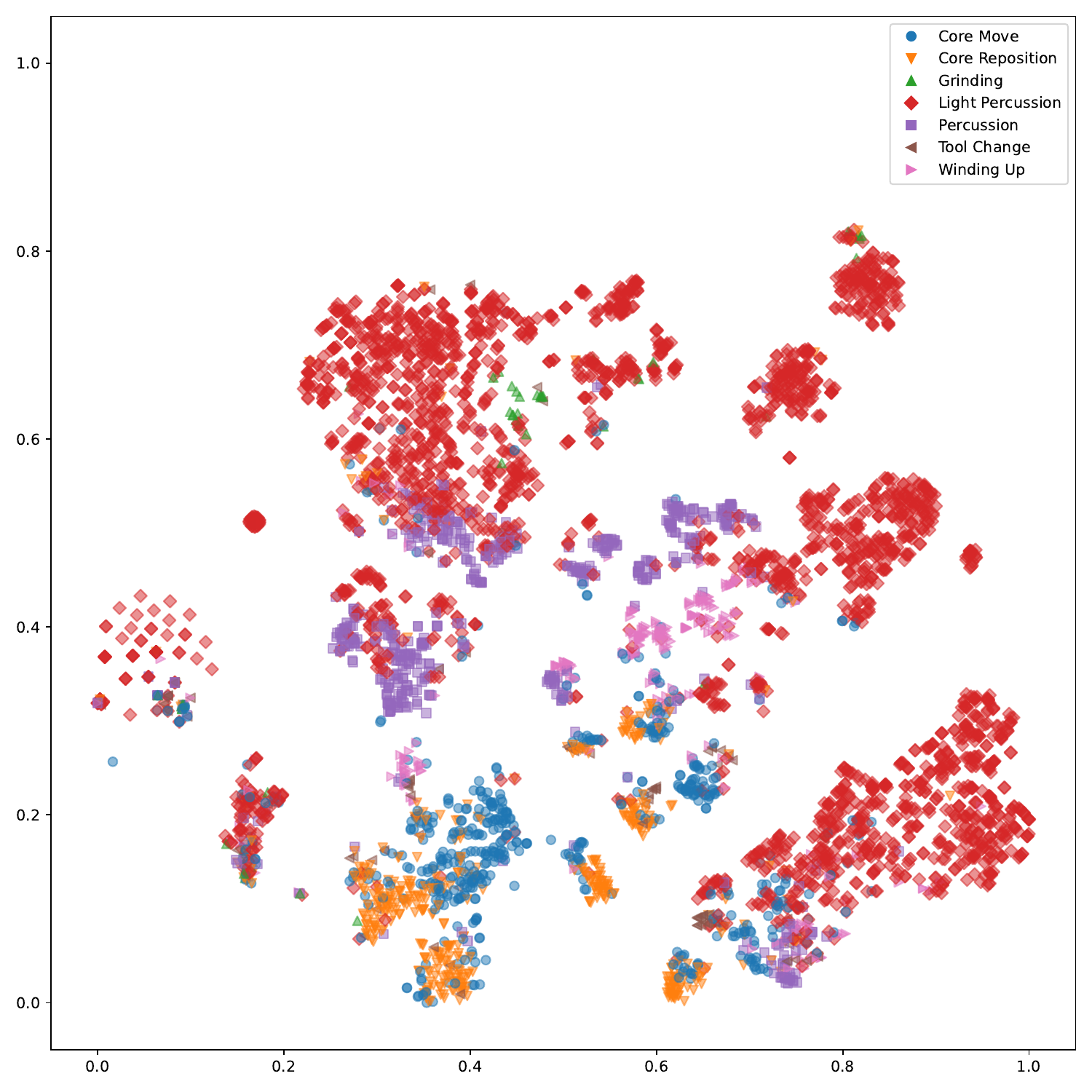}
\caption{TimesFormer model.}\label{tab:visual-timesformer}
\end{subfigure}
\begin{subfigure}{0.33\linewidth}
\includegraphics[width=\linewidth]{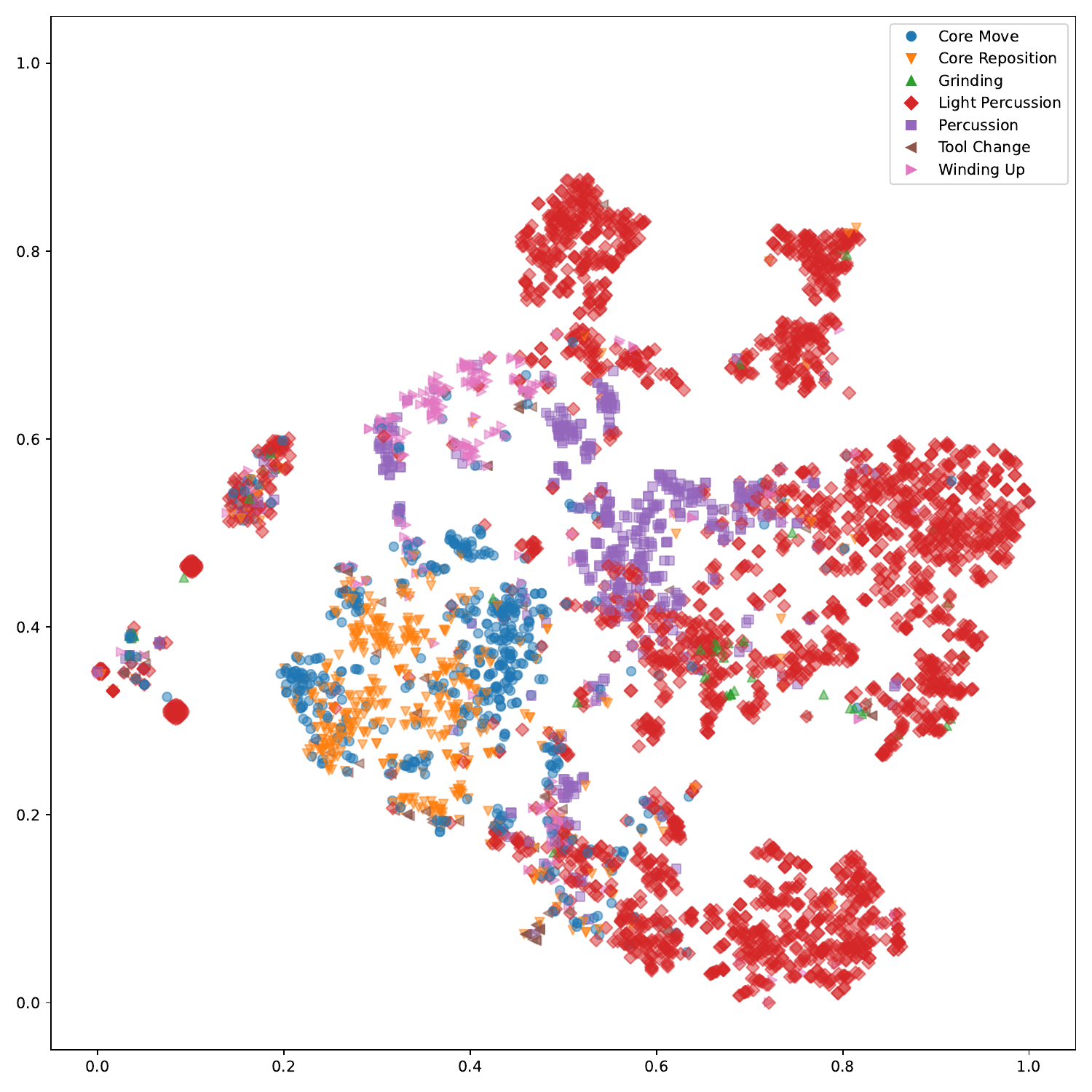}
\caption{ResNet+GRU model.}\label{visual:cm-resnet}
\end{subfigure}
\caption{Two-dimensional visualization on the embeddings learned from the trained models on HSTAG.} \label{table-visual}
\end{figure*}
In Table \ref{table-res}, the performance of all three trained models on the introduced HSTAG dataset are presented respectively. In general, the SOTA video classification models achieve at least $80\%$ overall classification accuracy. Among the SOTA models, the VideoMAE model outperforms the other two models in terms of overall $Acc$ and $F1_{mac}$. Nonetheless, the SOTA models yields a lower $F1_{mac}$ value and it implies the presence of imbalance class distribution as one of the most challenging aspects of the introduced domain-specific dataset. In regards to the class-specific performance, we observed that VideoMAE and TimeSformer achieve at least $60\%$ precision and recall on actions with more occurrence, which are Percussion, Light Percussion, Core Move, Core Reposition, and Winding Up. The performance of ResNet+GRU is significantly lower. We partially ascribed the superior performance of VideoMAE to its high masking ratio among the high-frequency action frames. This masking characteristic helps to extract a more descriptive and distinct representation from those actions. Moreover, as a data-efficient learner, VideoMAE can effectively boost the learning performance in action classes with a small sample size, which partially alleviates the imbalance class distribution issue. However, all three models do not achieve a very good performance in classifying the Tool Change and Grinding action. Tool change consists of several complex sequences of steps on switching between different tools and those steps may share similar characteristics with other action classes such as Winding Up, which makes it more challenging to distinguish. The use of different tools also increases the within-class variety. As the minority class, the Grinding action has much fewer number of samples than other action classes and the variety in different angle of views needs sufficient samples to learn its distinct characteristics.

Figure \ref{table-visual} provides a visualization of the sample distribution using the embedding learned from each SOTA model. We use the trained feature extractors, \textit{i.e.} the model with the last classification layer removed, to extract feature vectors from video clips in the test split video clips. We utilized the t-distributed Stochastic Neighbor Embedding (t-SNE) \cite{van2008visualizing} to reduce the feature vectors to 2-dimensional space for visualization. As shown in Figure \ref{table-visual}, action classes are more separable from each other in the embedding space of the VideoMAE model versus other SOTA models, which is consistent with the classification performance in Table \ref{table-res}.
\subsection{Analysis on class separability}
\begin{table}[t]
\begin{subtable}{\linewidth}\centering
\setlength{\tabcolsep}{3pt}
\begin{tabular}{cccc}\toprule
       Model                 & VideoMAEv2 & TimeSformer & ResNet+GRU \\ \midrule
$\bar{d}_{inter}$            & 12.02 & 27.95  & 18.56 \\
$\bar{d}_{intra}$            & 7.44 & 19.00 & 14.61 \\\midrule
$\bar{d}_{inter}/\bar{d}_{intra}$ & \textbf{1.62} & 1.47 & 1.27 \\\bottomrule
\end{tabular}
\caption{Inter-class and intra-class distances.}\label{table-distance}
\end{subtable}
\vspace{1pt}

\begin{subtable}{\linewidth}\centering
\setlength{\tabcolsep}{1pt}
\begin{tabular}{cccccccc}\toprule
\diagbox{$k_s$}{$k_t$} & \scriptsize Percussion & \begin{tabular}{c}\scriptsize Light\\ \scriptsize Percussion\end{tabular} & \begin{tabular}{c}\scriptsize Core\\ \scriptsize Move\end{tabular} & \begin{tabular}{c}\scriptsize Core\\ \scriptsize Reposition\end{tabular} & \scriptsize Grinding & \begin{tabular}{c}\scriptsize Tool\\ \scriptsize Change\end{tabular} & \begin{tabular}{c}\scriptsize Winding\\ \scriptsize Up\end{tabular} \\\midrule
\scriptsize Percussion       &0.711&\textbf{1.092}&1.246&1.346&1.279&1.258&1.162 \\
\scriptsize Light Percussion &     &0.710&1.180&1.222&\textbf{1.160}&1.153&\textbf{1.128} \\
\scriptsize Core Move        &     &     &0.898&\textbf{1.015}&1.185&1.134&1.193 \\
\scriptsize Core Reposition  &     &     &     &0.793&1.184&\textbf{1.102}&1.247 \\
\scriptsize Grinding         &     &     &     &     &0.879&1.176&1.232 \\
\scriptsize Tool Change      &     &     &     &     &     &1.042&1.180 \\
\scriptsize Winding Up       &&&&&&& 0.883 \\\bottomrule
\end{tabular}
\caption{Normalized cross-class distances $\hat{d}_{<k_s, k_t>}$ of VideoMAEv2.}\label{table-distance-pair-videomaeve}
\end{subtable}
\vspace{1pt}

\begin{subtable}{\linewidth}\centering
\setlength{\tabcolsep}{1pt}
\begin{tabular}{cccccccc}\toprule
\diagbox{$k_s$}{$k_t$} & \scriptsize Percussion & \begin{tabular}{c}\scriptsize Light\\ \scriptsize Percussion\end{tabular} & \begin{tabular}{c}\scriptsize Core\\ \scriptsize Move\end{tabular} & \begin{tabular}{c}\scriptsize Core\\ \scriptsize Reposition\end{tabular} & \scriptsize Grinding & \begin{tabular}{c}\scriptsize Tool\\ \scriptsize Change\end{tabular} & \begin{tabular}{c}\scriptsize Winding\\ \scriptsize Up\end{tabular} \\\midrule
\scriptsize Percussion       &0.911&\textbf{1.120}&1.192&1.320&1.211&1.195&1.103 \\
\scriptsize Light Percussion &     &0.757&1.126&1.221&0.930&\textbf{1.077}&1.072 \\
\scriptsize Core Move        &     &     &0.904&\textbf{0.967}&1.111&1.020&1.118 \\
\scriptsize Core Reposition  &     &     &     &0.859&1.131&1.043&1.217 \\
\scriptsize Grinding         &     &     &     &     &0.859&1.044&1.168 \\
\scriptsize Tool Change      &     &     &     &     &     &0.948&1.071 \\
\scriptsize Winding Up       &&&&&&& 0.864 \\\bottomrule
\end{tabular}
\caption{Normalized cross-class distances $\hat{d}_{<k_s, k_t>}$ of TimeSformer.}\label{table-distance-pair-timesformer}
\end{subtable}
\vspace{1pt}

\begin{subtable}{\linewidth}\centering
\setlength{\tabcolsep}{1pt}
\begin{tabular}{cccccccc}\toprule
\diagbox{$k_s$}{$k_t$} & \scriptsize Percussion & \begin{tabular}{c}\scriptsize Light\\ \scriptsize Percussion\end{tabular} & \begin{tabular}{c}\scriptsize Core\\ \scriptsize Move\end{tabular} & \begin{tabular}{c}\scriptsize Core\\ \scriptsize Reposition\end{tabular} & \scriptsize Grinding & \begin{tabular}{c}\scriptsize Tool\\ \scriptsize Change\end{tabular} & \begin{tabular}{c}\scriptsize Winding\\ \scriptsize Up\end{tabular} \\\midrule
\scriptsize Percussion       &0.874&\textbf{1.073}&1.101&1.133&1.065&1.059&1.060 \\
\scriptsize Light Percussion &     &0.853&1.122&1.113&\textbf{0.940}&1.046&1.100 \\
\scriptsize Core Move        &     &     &0.894&\textbf{0.904}&1.059&0.976&1.101 \\
\scriptsize Core Reposition  &     &     &     &0.814&1.036&\textbf{0.946}&1.106 \\
\scriptsize Grinding         &     &     &     &     &0.845&0.988&1.104 \\
\scriptsize Tool Change      &     &     &     &     &     &0.901&\textbf{1.042} \\
\scriptsize Winding Up       &&&&&&& 0.908 \\\bottomrule
\end{tabular}
\caption{Normalized cross-class distances $\hat{d}_{<k_s, k_t>}$ of ResNet+GRU.}\label{table-distance-pair-resnet}
\end{subtable}

\caption{Separability and cross-class correlation of the learned embedding of different models.}
\end{table}

We further investigate the separability and class correlation of the learned embedding qualitatively by calculating the average distance among the embedding vectors. For total of $N$ video clips, let $k_i$ be the action class label of the $i$-th clip, and $K$ is the total number of action classes. The extracted feature vector for each clip is $\mathbf{x}_{i}$. We calculate the average inter-class and intra-class Euclidean distances by
\begin{align}
\bar{d}_{inter}=\text{Average}_{i,j|k_i\neq k_j} ||\mathbf{x}_{i} - \mathbf{x}_{j}||_2, \\
\bar{d}_{intra}=\text{Average}_{i,j|k_i = k_j} ||\mathbf{x}_{i} - \mathbf{x}_{j}||_2.
\end{align}
We show the distances in Table \ref{table-distance}, and $\bar{d}_{inter}/\bar{d}_{intra}$ indicates how separable the classes are in the feature space. VideoMAE model also achieves the largest $\bar{d}_{inter}/\bar{d}_{intra}$, which supports the fact that the VideoMAE model learns a more distinct representation space among action classes.

The class correlation is measure by the average cross-class distance between pairs of feature vectors of two classes $k_s$ and $k_t$
\begin{equation}
\bar{d}_{<k_s, k_t>} = \text{Average}_{i,j|k_i = k_s, k_j=k_t} ||\mathbf{x}_{i} - \mathbf{x}_{j}||_2.
\end{equation}
For easier comparison among different models, we normalize the cross-class distance by the overall average distance
\begin{equation}
\hat{d}_{<k_s, k_t>} = \frac{\bar{d}_{<k_s, k_t>}}{\text{Average}_{i,j|i\neq j} ||\mathbf{x}_{i} - \mathbf{x}_{j}||_2}.
\end{equation}
The normalized distances are shown in Table \ref{table-distance-pair-videomaeve}, \ref{table-distance-pair-timesformer}, and \ref{table-distance-pair-resnet}. The minimum pairwise normalized cross-class distance is bold for each action. We can observe that the VideoMAEv2 model has a larger minimum pairwise cross-class distance than the other two models. This observation supports the fact the VideoMAEv2 
achieves better performance in classifying the highly similar action classes.

Overall, from our experimental analysis and comparison studies, several challenging aspects of HSTAG are observed and justified as follows: 
\begin{enumerate}
    \item High-frequency action frames: Percussion, Light Percussion, Core Move, and Core Reposition are high-frequency actions. This characteristic induced the temporal redundancy among video contents and caused poor learning performance.  
    \item High-similarity among different action classes: Among all seven classes, Percussion, Light Percussion, Core Move, Core Reposition, Tool Change, and Winding Up share similar characteristics, which made it more challenging to learn the distinct representations across highly similar action sequence.
    \item Presence of imbalanced class distributions: Actions including Grinding, Tool Change, and Winding Up have relatively smaller sample sizes compared to other actions. This phenomenon increased the difficulty in generalizing the model performance on all seven actions. 
\end{enumerate}

\section{Conclusion}
This paper presented a challenging domain-specific benchmark dataset for fine-grained human behavior recognition tasks, namely Human Stone Toolmaking Action Grammar (HSTAG). It consists of $18,739$ action videos with high-quality annotation information that maps them into seven point-event actions in stone tool-making activities. Unlike the existing human action recognition datasets, HSTAG are more challenging from several aspects: \begin{enumerate*}[label=(\roman*)]\item high-frequency action patterns with fine-grained human annotations; \item complexity and high-similarity among a sequence of actions; \item high imbalance ratio among action classes. \end{enumerate*} Using several SOTA temporal human action classification models, we conducted the experimental analysis and comparison study on HSTAG. Experimental results justified the challenging aspects of the HSTAG dataset. As a challenging dataset, HSTAG can serve as a valuable benchmark to evaluate the generalizability of the SOTA techniques for human action recognition in rarely seen domain. Furthermore, HSTAG can potentially advance future research in novel CV approaches on classifying temporal action sequences with unbalanced class distributions.

\section*{Acknowledgment}
This work is supported in part by the International Society of
Human Ethology’s Owen Aldis Award project titled “Sealed in stones: The computational ethology of stone toolmaking and its implications to the evolution of cultural transmission. (C.L.)”
\bibliographystyle{IEEEtran}
\bibliography{refers}

\end{document}